\documentclass[11pt,a4paper]{article}
 
\usepackage[utf8]{inputenc}
\usepackage[T1]{fontenc}
\usepackage{amsmath,amssymb,amsthm}
\usepackage{algorithm}
\usepackage{algorithmic}
\usepackage{graphicx}
\usepackage{booktabs}
\usepackage{multirow}
\usepackage{hyperref}
\usepackage{xcolor}
\usepackage{enumitem}
\usepackage{subcaption}
\usepackage{float}
\usepackage{geometry}
\usepackage{natbib}
\usepackage{thmtools}
\usepackage{mathtools}
\usepackage{bm}
\usepackage{url}
\usepackage{array}
\usepackage{longtable}
 
\hypersetup{
  colorlinks=true,
  linkcolor=blue,
  citecolor=blue,
  urlcolor=blue
}
 
\newtheorem{definition}{Definition}[section]
\newtheorem{remark}{Remark}[section]
 
\newcommand{\N}{\mathbb{N}}

\newcommand{\Prob}{\mathbb{P}}
\newcommand{\calX}{\mathcal{X}}
\newcommand{\calC}{\mathcal{C}}
\newcommand{\calF}{\mathcal{F}}
\newcommand{\calS}{\mathcal{S}}
\newcommand{\calH}{\mathcal{H}}
\newcommand{\calD}{\mathcal{D}}
\newcommand{\calM}{\mathcal{M}}
\newcommand{\calQ}{\mathcal{Q}}
\newcommand{\calB}{\mathcal{B}}
\newcommand{\calT}{\mathcal{T}}
\newcommand{\calA}{\mathcal{A}}
\newcommand{\bx}{\bm{x}}

\newcommand{\TBA}{\textsc{TBA}}
\newcommand{\TBATPE}{\textsc{TBA$\rightarrow$TPE}}
 
\title{
Feasible-First Exploration for Constrained ML Deployment Optimization\\
in Crash-Prone Hierarchical Search Spaces
}
 
\author{
Christian Lysenstøen\\
Department of Electrical Engineering and Computer Sciences\\
University of California, Berkeley\\
\texttt{christian@lysenstoen.net}
}
 
\date{April 2026}
 
\begin{document}
 
\maketitle
 
\begin{abstract}
Deploying machine learning models under production constraints requires joint optimization over model family, quantization scheme, runtime backend, and serving configuration. This induces a hierarchical mixed-variable search space in which many configurations are invalid: evaluations may crash, exceed memory limits, or violate latency constraints. Standard black-box optimizers such as Tree-structured Parzen Estimators (TPE) and constrained Bayesian optimization are effective when valid configurations are common, but they can spend a large fraction of a small evaluation budget on invalid or uninformative trials in hostile deployment spaces. This paper studies that regime and asks whether optimization should be decomposed into an explicit exploration stage followed by model-guided exploitation.
 
We propose \emph{Thermal Budget Annealing} (\TBA{}), a feasible-first exploration procedure that maps valid and feasible regions before warm-starting TPE. The method includes two robustness mechanisms for hostile hardware: \emph{trial timeouts} that abort clearly infeasible evaluations early, and \emph{subspace blacklisting} that temporarily suppresses categorical subspaces after repeated failures. We also introduce \emph{DeployBench}, a benchmark suite for deployment optimization with hierarchical structure, hidden crash zones, hard constraints, and unequal evaluation costs. On synthetic benchmarks and real GPU deployment with five pre-trained vision models across five GPU targets (NVIDIA H100, A100, RTX~5080, L4, and T4), the proposed hybrid improves model-family discovery under tight constraints while reducing wasted budget relative to cold-start TPE. The hybrid discovers the globally optimal model family (\texttt{vit\_tiny}) in 5/5 seeds on H100, 4/5 on A100 and L4, and 3/5 on T4, compared to TPE's 3/5 on H100, 4/5 on A100, 3/5 on L4, and 3/5 on T4. On the RTX~5080 with 10 seeds, the hybrid discovers \texttt{vit\_tiny} in 8/10 seeds versus TPE's 3/10. Multi-hardware experiments show that the hybrid matches or exceeds TPE on discovery across the tested hardware settings, with the clearest gains on H100 and RTX~5080. The results support a narrower but practically important conclusion: in partially invalid deployment spaces, early exploration quality determines which model families are available to the exploitation phase.
\end{abstract}
 
 
\section{Introduction}
\label{sec:introduction}
 
Modern ML deployment rarely consists of a single model-selection decision. Practical deployment requires choosing a model family, quantization scheme, runtime backend, batch size, and backend-specific execution parameters under strict constraints on latency, memory, and, in some settings, power. The resulting optimization problem is structurally different from conventional hyperparameter optimization. First, the search space is \emph{hierarchical}: the activeness of some parameters depends on earlier structural choices. Second, the space is often \emph{partially invalid}: some configurations fail outright because of out-of-memory events, unsupported backend--operator combinations, compiler failures, or quantization incompatibilities. Third, evaluation costs are unequal because exporting, compiling, and benchmarking different configurations can vary from seconds to minutes.
 
These characteristics matter most when the evaluation budget is small. A method that spends its first several trials on invalid configurations may never observe the model family that would have produced the best feasible deployment. This issue is especially acute for model-guided optimizers that rely on early samples to shape later exploitation. If early observations are narrow, then later sampling may reinforce that narrow view. The central question in this paper is therefore not whether a new optimizer can dominate established methods in all settings, but whether a \emph{feasible-first exploration stage} improves performance in deployment spaces that are crash-heavy, hierarchical, and budget-constrained.
 
This question is timely for two reasons. First, recent work in hyperparameter optimization and constrained optimization has emphasized hierarchical spaces, hidden constraints, and warm-starting as first-class concerns rather than secondary engineering details \citep{akiba2019optuna, awad2021dehb, lindauer2022smac3, watanabe2023ctpe, tian2024becbo}. Second, deployment optimization itself has become a systems problem: practical inference quality depends on the interaction between model architecture, quantization, runtime, and hardware target, not on model choice alone \citep{chen2018tvm, reddi2020mlperf, zaharia2024compound, agrawal2024vidur}.
 
The proposed method is simple. Phase~1 uses crash-aware simulated annealing to search for feasible regions and diversify model-family discovery. Phase~2 warm-starts Optuna's TPE sampler from the Phase~1 history. Two additional robustness mechanisms address hostile hardware environments: trial timeouts abort evaluations whose early latency measurements already exceed $5\times$ the constraint threshold, and subspace blacklisting temporarily suppresses categorical values (e.g., a specific quantization mode) after three consecutive failures. The resulting method, \TBATPE{}, is not presented as a replacement for TPE in benign spaces. The claim is narrower: when invalidity is high and budgets are small, explicit feasible-first exploration can improve model-family discovery and reduce wasted budget enough to make model-guided exploitation more effective.
 
\paragraph{Contributions.}
\begin{enumerate}[leftmargin=*]
\item We formalize budget-constrained deployment optimization in partially invalid hierarchical spaces and identify \emph{premature exploitation} as a central failure mode of model-guided search in crash-heavy regimes.
\item We propose \TBA{}, a feasible-first exploration procedure with adaptive temperature control, crash-aware state transitions, trial timeouts, subspace blacklisting, and an adaptive handoff into TPE.
\item We introduce \emph{DeployBench}, a benchmark suite for constrained deployment search with hidden crash regions, hierarchical conditionals, and unequal evaluation costs.
\item We show on synthetic tasks and real GPU deployment across five hardware targets that warm-starting TPE with feasible-first exploration improves discovery of the best model family under tight constraints while reducing wasted budget relative to cold-start TPE. The hybrid matches or exceeds TPE on discovery in the tested hardware settings, with the clearest gains on H100 (5/5 vs 3/5) and RTX~5080 (8/10 vs 3/10).
\end{enumerate}
 
\section{Background and Problem Setting}
\label{sec:background}
 
Deployment optimization differs from conventional hyperparameter optimization in ways that are operationally significant. In training-time HPO, a trial typically returns a scalar objective together with optional soft constraints \citep{hutter2019automated}. In deployment optimization, an evaluation may fail to return any useful objective at all due to runtime incompatibilities, export failures, out-of-memory conditions, or compiler exceptions. This creates a \emph{partially invalid} search problem in which optimization must reason simultaneously about objective quality, constraint satisfaction, and evaluability.
 
A second distinction is that deployment search spaces are naturally \emph{hierarchical}. Structural decisions such as backend selection activate or deactivate lower-level parameters. For example, selecting \texttt{torch\_compile} introduces compiler-related options that are irrelevant under eager execution, while selecting \texttt{onnxruntime} activates export and execution-provider choices. Consequently, the dimensionality of the active search space depends on the configuration itself.
 
A third distinction is that deployment trials have highly unequal cost. A configuration that runs in eager mode on a small model may complete in seconds, whereas a configuration that triggers graph export, compilation, or repeated failed allocation attempts may consume substantially more wall-clock time before yielding either a result or an exception. This cost asymmetry means that wasted evaluations are not merely statistically inefficient; they are operationally expensive.
 
These observations motivate a decomposition of the optimization process into two conceptual tasks: \emph{region discovery} and \emph{within-region optimization}. Region discovery concerns identifying model families and system configurations that can be evaluated successfully and satisfy constraints. Within-region optimization concerns improving the deployment objective after such regions have been identified. The central premise of this work is that standard model-guided optimizers conflate these tasks too early in crash-heavy settings.
 
\section{Related Work}
\label{sec:related}
 
\subsection{Hyperparameter Optimization}
 
The field of hyperparameter optimization has matured substantially over the past decade. Random search \citep{bergstra2012random} established that uniform sampling can outperform grid search in high-dimensional spaces. Bayesian optimization with Gaussian processes \citep{snoek2012practical, garnett2023bayesian} introduced model-guided sequential search, and Tree-structured Parzen Estimators \citep{bergstra2011algorithms} provided a scalable alternative for conditional spaces. Multi-fidelity methods such as Hyperband \citep{li2017hyperband} and BOHB \citep{falkner2018bohb} reduce cost by truncating unpromising trials, while DEHB \citep{awad2021dehb} combines Hyperband with evolutionary search for improved scalability. Optuna \citep{akiba2019optuna} and SMAC3 \citep{lindauer2022smac3} provide practical frameworks with support for conditional search spaces, and the recent c-TPE extension \citep{watanabe2023ctpe} adds native constraint-aware sampling to the TPE family. Existing HPO benchmarks \citep{eggensperger2021hpobench, pfisterer2022yahpo} have standardized evaluation methodology but generally assume that most configurations are evaluable.
 
These methods are designed for spaces where the vast majority of configurations produce valid results. The deployment setting studied here, in which 30--80\% of random configurations may crash or violate constraints, lies outside those assumptions.
 
\subsection{Constrained and Hidden-Constraint Optimization}
 
Constrained Bayesian optimization has been studied through probability-of-feasibility weighting \citep{gardner2014bayesian}, unknown-constraint modeling \citep{gelbart2014bayesian}, and information-based acquisition strategies \citep{hernandezlobato2016predictive}. More recently, optimization under hidden or partially observed constraints has received renewed attention \citep{tian2024becbo}. The present work connects that perspective to ML deployment, where crashes are not merely noisy evaluations but an intrinsic feature of the search space.
 
\subsection{ML Systems Optimization}
 
Deployment optimization for ML systems has gained attention through compiler-level approaches such as TVM \citep{chen2018tvm} and Ansor \citep{zheng2020ansor}, which automate operator scheduling. TensorRT and ONNX Runtime provide backend-specific runtime optimization. MLPerf \citep{reddi2020mlperf} has established the importance of inference benchmarking under realistic latency constraints. More recent systems work has examined edge deployment \citep{hymel2023edgeimpulse}, large-scale inference simulation \citep{agrawal2024vidur}, and the broader shift toward compound AI systems where deployment configuration spans multiple interacting components \citep{zaharia2024compound}. The problem considered here differs from this literature by treating deployment configuration itself as a constrained search problem over model family, quantization mode, backend, and serving parameters.
 
\subsection{Simulated Annealing and Hybrid Methods}
 
Simulated annealing \citep{kirkpatrick1983optimization} is a classical metaheuristic for combinatorial optimization. Adaptive temperature schedules and reheating strategies are well studied \citep{ingber1989very}. Evolutionary strategies \citep{hansen2001completely} provide an alternative paradigm for black-box search in mixed spaces. The contribution of the present work is not a new simulated annealing theory, but the use of feasible-first annealing as an exploration front-end for model-guided optimization in crash-heavy deployment spaces, augmented with robustness mechanisms that address the specific failure modes of hostile hardware environments.
 
\section{Problem Formalization}
\label{sec:problem}
 
\subsection{Hierarchical Deployment Search Space}
 
\begin{definition}[Hierarchical mixed-variable search space]
\label{def:search_space}
A hierarchical mixed-variable search space $\calX$ is defined by variables $\{x_1, \ldots, x_d\}$ where each variable has a type $\tau_i \in \{\text{categorical}, \text{integer}, \text{continuous}\}$, a domain $\calD_i$, and an optional activation condition $\phi_i: \calX \to \{0, 1\}$. A configuration $\bx \in \calX$ assigns values only to active variables. The active set $\calA(\bx) = \{i : \phi_i(\bx) = 1\}$ depends on the configuration itself.
\end{definition}
 
In the deployment setting studied here,
\begin{equation}
\calX = \calM \times \calQ \times \calB \times \calS \times \prod_{i \in \calA(\bx)} \calD_i,
\end{equation}
where $\calM = \{\texttt{resnet18}, \texttt{resnet50}, \texttt{mobilenet\_v2}, \texttt{efficientnet\_b0}, \texttt{vit\_tiny}\}$, $\calQ = \{\texttt{fp32}, \texttt{fp16}, \texttt{int8\_dynamic}\}$, $\calB = \{\texttt{pytorch\_eager}, \texttt{torch\_compile}, \texttt{onnxruntime}\}$, and $\calS = \{1, 2, 4, 8, 16, 32\}$.
 
\subsection{Constraints and Partial Invalidity}
 
The deployment optimization problem is:
\begin{equation}
\label{eq:deployment_problem}
\begin{aligned}
\bx^* = \arg\max_{\bx \in \calX} \quad & f(\bx) \\
\text{subject to} \quad & g_j(\bx) \leq c_j, \quad j = 1, \ldots, J, \\
& \bx \in \calX_{\text{valid}} \subseteq \calX.
\end{aligned}
\end{equation}
Here $f$ denotes a deployment objective such as accuracy or throughput, $g_j$ are constraint functions such as $\text{latency}_{p95}$ or peak memory, and $\calX_{\text{valid}}$ denotes the unknown subset where evaluation does not crash.
 
Total constraint violation is:
\begin{equation}
\label{eq:violation}
V(\bx) = \sum_{j=1}^{J}\max\left(0,\; g_j(\bx)-c_j\right).
\end{equation}
 
A configuration is feasible if $V(\bx)=0$ and $\bx \in \calX_{\text{valid}}$.
 
\begin{definition}[Crash zone]
\label{def:crash_zone}
The crash zone $\calX_{\text{crash}} \subset \calX$ is the subset of configurations for which evaluation fails to produce a valid result. The crash zone is not known a priori and must be discovered through evaluation.
\end{definition}
 
The effective search space decomposes as
\begin{equation}
\calX = \underbrace{\calX_{\text{crash}}}_{\text{crashed}} \cup
\underbrace{(\calX_{\text{valid}} \setminus \calC)}_{\text{valid but infeasible}} \cup
\underbrace{\calF}_{\text{feasible}},
\end{equation}
where $\calC = \{\bx : g_j(\bx)\le c_j,\; \forall j\}$ and $\calF = \calX_{\text{valid}}\cap \calC$.
 
\begin{definition}[Budget-constrained deployment optimization]
\label{def:budgeted}
Given $\calX$, objective $f$, constraints $\{(g_j,c_j)\}_{j=1}^{J}$, and evaluation budget $B \in \N$, the task is to find the best feasible configuration using at most $B$ evaluations, where each evaluation may crash and has variable cost $t(\bx)>0$.
\end{definition}
 
Two metrics capture optimizer performance. The first is simple regret:
\begin{equation}
\label{eq:regret}
r(n) = f(\bx^*_{\text{global}}) - \max_{i\le n,\;\bx_i\in\calF} f(\bx_i).
\end{equation}
The second is wasted budget fraction:
\begin{equation}
\label{eq:wasted_budget}
W(n) = \frac{\left|\left\{i \le n : \bx_i \not\in \calF\right\}\right|}{n}.
\end{equation}
 
\subsection{A Heuristic Characterization of Premature Exploitation}
 
The central failure mode considered in this work is \emph{premature exploitation}. In a crash-heavy space, an optimizer that transitions into model-guided sampling after only a small number of successful evaluations may have observed only a narrow subset of model families. Subsequent exploitation then reinforces that biased sample.
 
\begin{remark}[Probabilistic intuition for premature exploitation]
\label{rem:premature}
If the probability of sampling a specific model family in one trial is $p_m \approx 1/K$ for $K$ families, and the survival probability is $(1-\rho_{\text{crash}})$, then the probability of discovering that family in $n_0$ random trials is
\begin{equation}
\label{eq:discovery_prob}
\Prob(\text{discover family } m \text{ in } n_0 \text{ trials}) = 1 - \left(1 - p_m(1-\rho_{\text{crash}})\right)^{n_0}.
\end{equation}
For $K=5$, $\rho_{\text{crash}}=0.6$, and $n_0=10$, this yields approximately $0.57$, which predicts that only about 4--6 of 10 seeds would discover a given family during the initial exploration stage. This is directionally consistent with the observed behavior of cold-start TPE in the edge-tight setting (Section~\ref{sec:results}).
\end{remark}
 
\section{Research Hypotheses}
\label{sec:hypotheses}
 
The proposed method is motivated by three hypotheses.
 
\paragraph{H1: Early structural coverage matters.}
In hierarchical deployment spaces, the main risk of early optimization is not merely poor local refinement but insufficient structural coverage. If early evaluations fail to touch promising model families or backend regimes, later model-guided search is conditioned on a biased subset of the space.
 
\paragraph{H2: Invalidity changes the exploration--exploitation trade-off.}
In standard black-box optimization, random or quasi-random early trials are often adequate to initialize a surrogate. In partially invalid spaces, the same number of trials may produce too few successful observations to support reliable model-guided exploitation.
 
\paragraph{H3: Warm-started exploitation is preferable to cold-start exploitation in hostile spaces.}
If an exploration phase can deliberately accumulate a more diverse set of valid and feasible observations, then a subsequent model-guided optimizer should perform better than if it had spent the same budget on unguided early exploration.
 
These hypotheses define the scientific scope of the paper. The contribution is therefore not framed as a universally superior optimizer, but as evidence for a specific architectural claim: in crash-heavy constrained deployment spaces, explicit feasible-first exploration improves the quality of subsequent model-guided search.
 
\section{Method: Thermal Budget Annealing}
\label{sec:method}
 
\subsection{Overview}
 
\TBATPE{} consists of two phases. Phase~1 uses simulated annealing to search for feasible regions while deliberately traversing structural choices. Phase~2 injects the full history from Phase~1 into a constrained TPE study and uses TPE for model-guided exploitation. Two additional robustness mechanisms---trial timeouts and subspace blacklisting---reduce wasted budget on hostile hardware by aborting clearly doomed trials and avoiding repeatedly failing categorical subspaces. The full optimizer is summarized in Algorithm~\ref{alg:tba_tpe}.
 
\begin{algorithm}[t]
\caption{\TBATPE{} Hybrid Optimizer}
\label{alg:tba_tpe}
\begin{algorithmic}[1]
\REQUIRE Search space $\calX$, constraints $\{(g_j, c_j)\}$, objective $f$, budget $B$
\STATE $\calH \leftarrow \emptyset$, $T \leftarrow T_0$, phase $\leftarrow$ \texttt{feasibility}
\STATE Initialize subspace tracker $\calT$ and early-stopping threshold $\tau = 5 \cdot c_{\text{latency}}$
\STATE Sample $n_{\text{init}}$ random configurations; evaluate and add to $\calH$
\STATE $\bx_{\text{curr}} \leftarrow$ best feasible in $\calH$ (or least-violating if none feasible)
\FOR{$n = n_{\text{init}} + 1$ to adaptive handoff}
    \STATE Compute structural mutation probability $p_s(n)$
    \STATE Mask blacklisted categorical values from $\calT$ when proposing
    \IF{elite restart condition met and $n < 0.7B$}
        \STATE $\bx' \leftarrow$ 2nd or 3rd best feasible from $\calH$
    \ELSE
        \STATE $\bx' \leftarrow \text{ProposeNeighbor}(\bx_{\text{curr}}, T, p_s(n))$
    \ENDIF
    \STATE Evaluate $\bx'$ with early stopping: abort if latency $> \tau$ after warmup
    \STATE Add $(\bx', r')$ to $\calH$; update $\calT$ with outcome
    \IF{$r'$ indicates crash or early stop}
        \STATE Update $T$ as a non-improving step; keep $\bx_{\text{curr}}$
    \ELSIF{phase = \texttt{feasibility}}
        \STATE Apply violation-based Metropolis update
    \ELSE
        \STATE Apply objective-based Metropolis update with hard feasibility filter
    \ENDIF
    \IF{HandoffReady($\calH$)}
        \STATE \textbf{break}
    \ENDIF
\ENDFOR
\STATE Create Optuna study with constrained TPE sampler \citep{watanabe2023ctpe}
\FOR{each $(\bx_i, r_i) \in \calH$}
    \STATE Inject as completed trial with objective value and constraint metadata
\ENDFOR
\FOR{$n = |\calH| + 1$ to $B$}
    \STATE Request next configuration from TPE (full search space, no blacklisting)
    \STATE Evaluate with early stopping and report result to the study
\ENDFOR
\RETURN best feasible configuration observed
\end{algorithmic}
\end{algorithm}
 
\subsection{Phase 1: Feasible-First Simulated Annealing}
 
Phase~1 operates with a two-stage objective. Before any feasible configuration has been found, the optimizer minimizes total violation $V(\bx)$. Once a feasible configuration is found, the objective switches to maximizing $f(\bx)$ over feasible states.
 
The temperature follows an adaptive schedule with reheating:
\begin{equation}
\label{eq:temperature}
T_{n+1} =
\begin{cases}
\alpha_s T_n & \text{if improved},\\
T_n & \text{if not improved and patience not exceeded},\\
\min(\beta T_n,\; \gamma T_0) & \text{if patience exceeded}.
\end{cases}
\end{equation}
In the reported experiments, $\alpha_s \in \{0.92, 0.95\}$ with slower cooling during the feasibility stage to maintain broader exploration, $\beta=2.5$, $\gamma=0.8$, and patience is 4 consecutive non-improving trials.
 
\subsection{Acceptance Criteria}
 
The acceptance rule depends on the current stage.
 
\paragraph{Feasibility stage.}
A new state is accepted deterministically if it reduces total violation. Otherwise, it is accepted with probability
\begin{equation}
\Prob(\text{accept}) = \exp\left(-\frac{V(\bx')-V(\bx_{\text{curr}})}{T_n \cdot \sigma_V}\right),
\end{equation}
where
\begin{equation}
\sigma_V = \max\left(0.1,\; \frac{1}{J}\sum_{j=1}^{J}|c_j|\cdot 0.1\right).
\end{equation}
 
\paragraph{Optimization stage.}
If the proposal is infeasible, it is rejected outright. If it improves the objective, it is accepted. Otherwise, it is accepted with probability
\begin{equation}
\Prob(\text{accept}) = \exp\left(-\frac{f(\bx_{\text{curr}})-f(\bx')}{T_n \cdot \sigma_f}\right),
\end{equation}
where
\begin{equation}
\sigma_f = \max\left(0.01,\; \frac{f_{\max}-f_{\min}}{2}\right).
\end{equation}
 
\subsection{Hierarchical Neighbor Proposals}
 
Neighborhood proposals must respect the hierarchical structure of the space. With probability $p_s(n)$, the optimizer performs a structural move by changing a categorical variable such as model family, backend, or quantization mode. Conditional variables activated by the new structure are sampled from their valid domains; deactivated variables are removed. With probability $1-p_s(n)$, the optimizer performs a numerical move by perturbing an active integer or continuous parameter.
 
The structural mutation probability decays linearly over the budget:
\begin{equation}
\label{eq:p_structural}
p_s(n) = 0.5 - 0.35 \cdot \frac{n - n_{\text{init}}}{B - n_{\text{init}}}.
\end{equation}
This schedule reflects the intuition that early search should privilege structural exploration while later search should privilege local refinement.
 
\subsection{Crash Handling and Exploration Mechanisms}
 
Crashed configurations are never accepted as the current state. However, they are retained in history and therefore contribute to the optimizer's implicit map of invalid regions. This design allows the search to learn crash boundaries without accumulating in crash-prone states.
 
Two additional mechanisms stabilize exploration. \emph{Elite restart} replaces the current state with the 2nd or 3rd best feasible configuration every 12 trials during the first 70\% of budget. \emph{Decaying snap-back} returns to the global best feasible state with probability $0.4(1-n/B)$ after accepting a worse feasible move.
 
\subsection{Trial Timeouts}
\label{sec:trial_timeouts}
 
On hostile hardware such as the T4, certain configurations (e.g., \texttt{int8\_dynamic} quantization) produce latencies of $500$ms or more---$25\times$ the $20$ms constraint. Running these trials to completion wastes both wall-clock time and evaluation budget. Trial timeouts address this by aborting evaluations early when they are clearly infeasible.
 
After the initial warmup iterations, the profiler measures the average latency over the first few timed iterations. If this average exceeds $\tau = 5 \cdot c_{\text{latency}}$ (i.e., $100$ms for a $20$ms constraint), the trial is immediately aborted and marked as early-stopped. The factor of $5\times$ ensures that only \emph{catastrophically} infeasible trials are terminated; borderline-infeasible trials still run to completion and provide useful constraint information.
 
This mechanism is applied at the benchmark level and therefore benefits \emph{all} optimizers equally. In the reported experiments, trial timeouts saved 7--47 seconds per seed on the T4 and 6--25 seconds per seed on the H100, with the savings concentrated on \texttt{int8\_dynamic} and large-batch configurations.
 
\subsection{Subspace Blacklisting}
\label{sec:subspace_blacklisting}
 
A more targeted mechanism addresses the \emph{exploration trap} identified in early T4 experiments, where the SA phase repeatedly proposed configurations from a catastrophically infeasible categorical subspace (e.g., \texttt{quantization=int8\_dynamic}) without learning to avoid it.
 
The subspace tracker maintains a per-value failure counter for each categorical variable. When a categorical value accumulates three consecutive crash or infeasible outcomes, it is temporarily \emph{blacklisted}: the SA neighbor proposal masks that value from the categorical domain. Three safeguards prevent over-restriction:
\begin{enumerate}[leftmargin=*]
\item \emph{Cooldown}: after 8 trials, a blacklisted value is automatically restored to the active domain.
\item \emph{Success reset}: any successful trial using a blacklisted value immediately removes the blacklist.
\item \emph{Safety valve}: the tracker never blacklists \emph{all} values of any categorical variable; at least one value always remains available.
\end{enumerate}
 
Subspace blacklisting is applied only during Phase~1 (SA exploration). Phase~2 (TPE exploitation) receives the full, unrestricted search space. This design reflects the intent that blacklisting is an exploration-efficiency mechanism, not a permanent constraint on the search.
 
\subsection{Adaptive Handoff into TPE}
 
The handoff from exploration to exploitation occurs when the optimizer has observed sufficient evidence to initialize TPE reliably:
\begin{equation}
\label{eq:handoff}
\text{handoff if } n \ge n_{\min}, \quad N_{\text{feas}} \ge 5,\quad N_{\text{bad}} \ge 3,
\end{equation}
with a hard maximum of $n_{\max}=15$ trials. Here $N_{\text{feas}}$ is the number of feasible observations and $N_{\text{bad}}$ is the number of crashed or infeasible observations. This criterion ensures that the TPE phase receives both positive and negative information about the space.
 
\subsection{Phase 2: Warm-Started TPE}
 
All Phase~1 evaluations are injected into an Optuna study as completed trials, including crashed and infeasible configurations together with constraint metadata. The constrained TPE acquisition can be expressed schematically as
\begin{equation}
\label{eq:constrained_tpe}
\text{EI}_c(\bx) = \frac{\ell(\bx)}{g(\bx)} \cdot \Prob(\bx \in \calC \mid \calH),
\end{equation}
where $\ell(\bx)$ and $g(\bx)$ are density estimates over good and bad observations, respectively. The crucial point is that Phase~2 begins from a broader and more informative set of initial observations than cold-start TPE would obtain from unguided early trials.
 
\subsection{Computational Cost}
 
The practical cost of deployment optimization is determined jointly by trial count and wall-clock time. Let $t(\bx)$ denote the time required to evaluate configuration $\bx$. Then cumulative wall-clock cost is
\begin{equation}
T_{\text{wall}}(n) = \sum_{i=1}^{n} t(\bx_i).
\end{equation}
Trial timeouts directly reduce $T_{\text{wall}}$ by aborting catastrophically infeasible evaluations before completion. Subspace blacklisting indirectly reduces it by steering proposals away from subspaces where $t(\bx)$ is high and outcomes are consistently infeasible.
 
\section{DeployBench: A Benchmark for Crash-Heavy Deployment Optimization}
\label{sec:benchmark}
 
\subsection{Design Principles}
 
Existing HPO benchmarks \citep{eggensperger2021hpobench, pfisterer2022yahpo} assume that most configurations produce valid results. DeployBench is designed to reflect deployment regimes where this assumption fails:
\begin{enumerate}[leftmargin=*]
\item many random configurations are invalid (30--80\% invalidity),
\item the space is hierarchical and mixed-variable,
\item constraints are hard rather than softly penalized,
\item evaluation costs vary substantially,
\item each trial is logged in JSONL format for reproducible behavioral analysis.
\end{enumerate}
 
\subsection{Synthetic Benchmarks}
 
Two synthetic tasks were used during development and evaluation. \emph{Crashy Branin} augments the standard Branin function with a categorical mode variable, an integer resolution variable, explicit crash zones, and feasibility constraints, yielding approximately 65\% combined invalidity. \emph{Hierarchical Rosenbrock} activates different subsets of continuous variables according to a categorical mode, thereby varying effective dimensionality across the space.
 
\subsection{Real Deployment Benchmark}
 
The real benchmark evaluates deployment configurations on ImageNette \citep{howard2020imagenette} using five pretrained vision models, three backends, three quantization modes, six batch sizes, and a conditional thread parameter. The resulting search space is summarized in Table~\ref{tab:search_space}.
 
\begin{table}[t]
\centering
\caption{Deployment search space variables.}
\label{tab:search_space}
\begin{tabular}{lllc}
\toprule
\textbf{Variable} & \textbf{Type} & \textbf{Domain} & \textbf{Conditional} \\
\midrule
\texttt{model\_name} & Categorical & \{resnet18, resnet50, mobilenet\_v2, efficientnet\_b0, vit\_tiny\} & No \\
\texttt{backend} & Categorical & \{pytorch\_eager, torch\_compile, onnxruntime\} & No \\
\texttt{quantization} & Categorical & \{fp32, fp16, int8\_dynamic\} & No \\
\texttt{batch\_size} & Integer & $\{1,2,4,8,16,32\}$ & No \\
\texttt{num\_threads} & Integer & $[1,8]$ & backend $\neq$ \texttt{torch\_compile} \\
\bottomrule
\end{tabular}
\end{table}
 
Before accounting for conditional inactivity, this space contains 2160 structural combinations. Known crash patterns include \texttt{onnxruntime} with various model--precision combinations, \texttt{int8\_dynamic} producing extreme latencies on slow GPUs, and large model--batch size combinations causing GPU OOM. These patterns are intentionally not filtered from the search space.
 
Each evaluation follows a standardized procedure: model loading, optional quantization, backend compilation or export, 30 warmup iterations, 100 timed iterations (with trial timeout check after warmup), resource measurement, accuracy evaluation on 500 ImageNette validation images, and explicit memory cleanup. All evaluations are wrapped in exception handling so that a crash produces a logged result rather than terminating the experiment.
 
\begin{table}[t]
\centering
\caption{Deployment scenarios.}
\label{tab:scenarios}
\begin{tabular}{lllc}
\toprule
\textbf{Scenario} & \textbf{Constraints} & \textbf{Objective} & \textbf{Budget} \\
\midrule
\texttt{edge\_tight} & latency$_{p95}\le 20$ms, memory $\le 512$MB & maximize accuracy & 25 \\
\texttt{server\_throughput} & latency$_{p99}\le 100$ms, memory $\le 4096$MB & maximize throughput & 25 \\
\bottomrule
\end{tabular}
\end{table}
 
\section{Experimental Setup}
\label{sec:setup}
 
\subsection{Baselines}
 
Three baselines are evaluated under identical search spaces, budgets, and seeds:
\begin{enumerate}[leftmargin=*]
\item \textbf{Random Search} \citep{bergstra2012random}: uniform random sampling, reporting the best feasible result.
\item \textbf{Optuna TPE} \citep{akiba2019optuna, watanabe2023ctpe}: constrained tree-structured Parzen estimation with native conditional handling.
\item \textbf{TBA (pure SA)}: the feasible-first exploration stage without TPE handoff. This ablation isolates the contribution of warm-starting.
\end{enumerate}
 
All optimizers share the same benchmark infrastructure, including trial timeouts. Subspace blacklisting is applied only within the TBA and \TBATPE{} optimizers during Phase~1.
 
\subsection{Metrics}
 
Performance is assessed along four dimensions:
\begin{enumerate}[leftmargin=*]
\item best feasible objective (mean $\pm$ std across seeds),
\item wasted budget fraction $W(B)$,
\item time to first feasible configuration,
\item model-family discovery rate.
\end{enumerate}
 
\subsection{Hardware, Software, and Protocol}
 
Primary experiments were conducted on an NVIDIA RTX~5080 Laptop GPU (16~GB GDDR7) with an Intel Core Ultra~9 275HX and 16~GB DDR5 RAM. Multi-hardware generalization experiments (Section~\ref{sec:multi_hw}) were conducted on four additional GPU targets via Google Colab: NVIDIA H100~80GB, A100~40GB, L4~24GB, and T4~16GB. Software versions: PyTorch~2.6, torchvision~0.21, ONNX Runtime~1.21, Optuna~4.2, Python~3.12. All primary experiments use 10 random seeds (0--9); multi-hardware experiments use 5 seeds (0--4). All experiments across all GPUs use the same codebase (v2-improvements branch, commit \texttt{7ea93e2}). Conditional parameters are exposed consistently across methods, and crashed trials are recorded rather than silently filtered.
 
\section{Results}
\label{sec:results}
 
\subsection{Synthetic Benchmarks: Budget Sweep}
 
Table~\ref{tab:budget_sweep} reports results on Crashy Branin across budgets from 10 to 50.
 
\begin{table}[t]
\centering
\caption{Crashy Branin: best feasible objective (higher is better) and wasted budget fraction. Bold indicates best objective at each budget.}
\label{tab:budget_sweep}
\begin{tabular}{c|cc|cc|cc|cc}
\toprule
& \multicolumn{2}{c|}{\TBATPE{}} & \multicolumn{2}{c|}{TBA (SA)} & \multicolumn{2}{c|}{TPE} & \multicolumn{2}{c}{Random} \\
$B$ & Obj & $W$ & Obj & $W$ & Obj & $W$ & Obj & $W$ \\
\midrule
10 & $-5.83$ & 34\% & $\mathbf{-5.65}$ & 40\% & $-7.29$ & 54\% & $-8.44$ & 65\% \\
15 & $-5.54$ & 25\% & $-4.71$ & 32\% & $\mathbf{-2.59}$ & 42\% & $-6.37$ & 66\% \\
20 & $\mathbf{-3.18}$ & 30\% & $-4.62$ & 27\% & $-4.19$ & 40\% & $-5.17$ & 63\% \\
25 & $-2.29$ & 28\% & $-3.68$ & 28\% & $\mathbf{-2.14}$ & 35\% & $-4.12$ & 64\% \\
30 & $-2.05$ & 30\% & $-2.93$ & 27\% & $\mathbf{-1.97}$ & 31\% & $-4.41$ & 67\% \\
40 & $-1.57$ & 26\% & $-3.95$ & 24\% & $\mathbf{-1.38}$ & 30\% & $-2.98$ & 63\% \\
50 & $-1.45$ & 23\% & $-2.66$ & 20\% & $\mathbf{-1.15}$ & 29\% & $-2.56$ & 66\% \\
\bottomrule
\end{tabular}
\end{table}
 
Three findings emerge from Table~\ref{tab:budget_sweep}. First, at $B=20$, the hybrid outperforms all baselines including TPE ($-3.18$ vs $-4.19$), indicating that warm-starting with crash-aware exploration data can outperform cold-start model-guided search when the budget is tight. Second, at $B=25$--$30$, the hybrid and TPE achieve comparable objectives, but the hybrid exhibits lower variance ($\pm 0.79$ vs $\pm 1.33$ at $B=25$), indicating more consistent performance. Third, the wasted budget fraction for the hybrid (23--34\%) is consistently lower than TPE's (29--54\%), with the largest gap at small budgets. This pattern is consistent with H2 and H3: the main gain comes from better early exploration rather than stronger asymptotic exploitation.
 
\begin{figure}[t]
\centering
\includegraphics[width=0.85\textwidth]{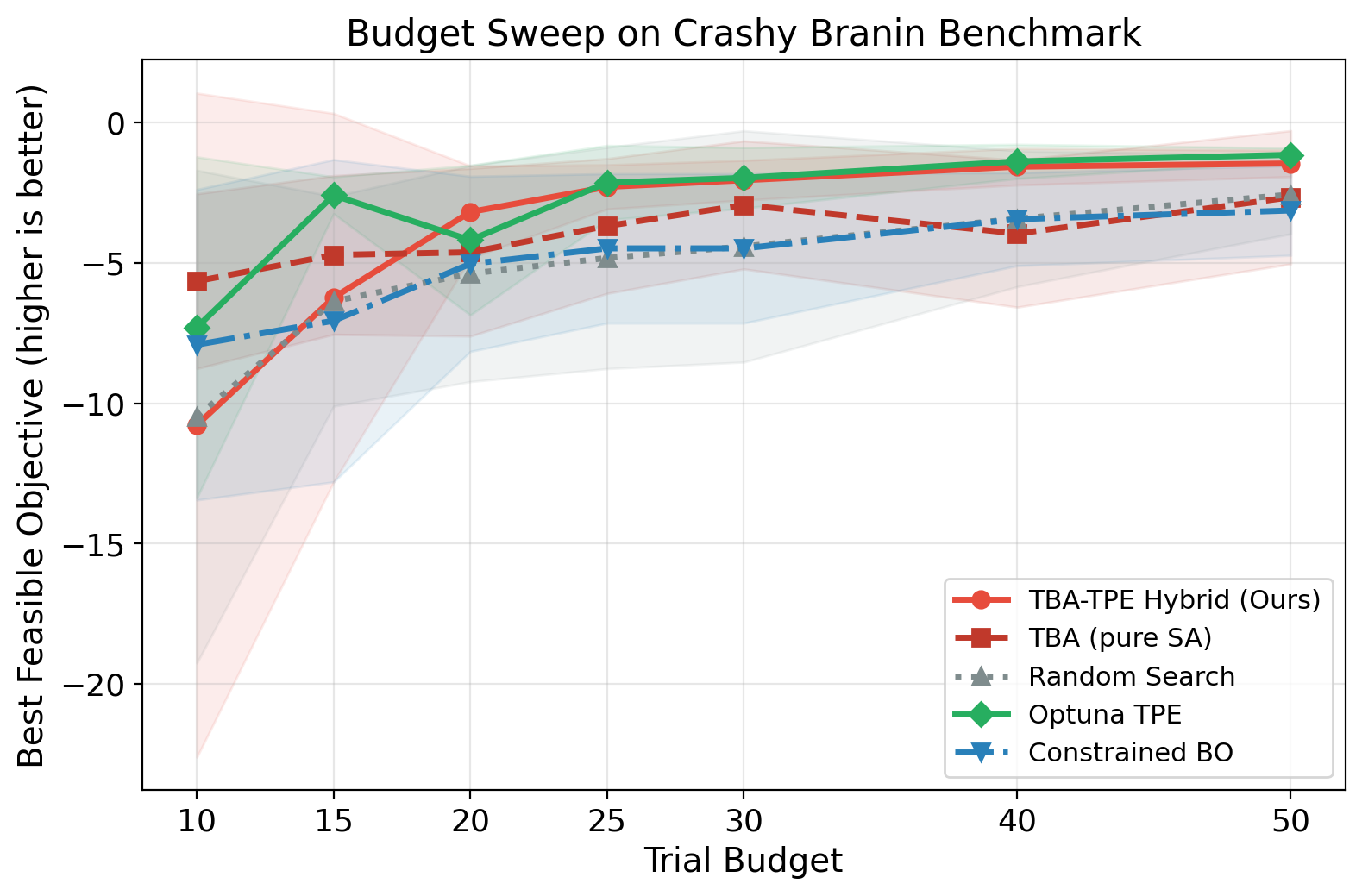}
\caption{Best feasible objective versus evaluation budget on Crashy Branin (10 seeds, mean $\pm$ std). The hybrid outperforms cold-start TPE at $B=20$ and tracks it closely at higher budgets, while pure SA and random search lag behind.}
\label{fig:budget_obj}
\end{figure}
 
\begin{figure}[t]
\centering
\includegraphics[width=0.85\textwidth]{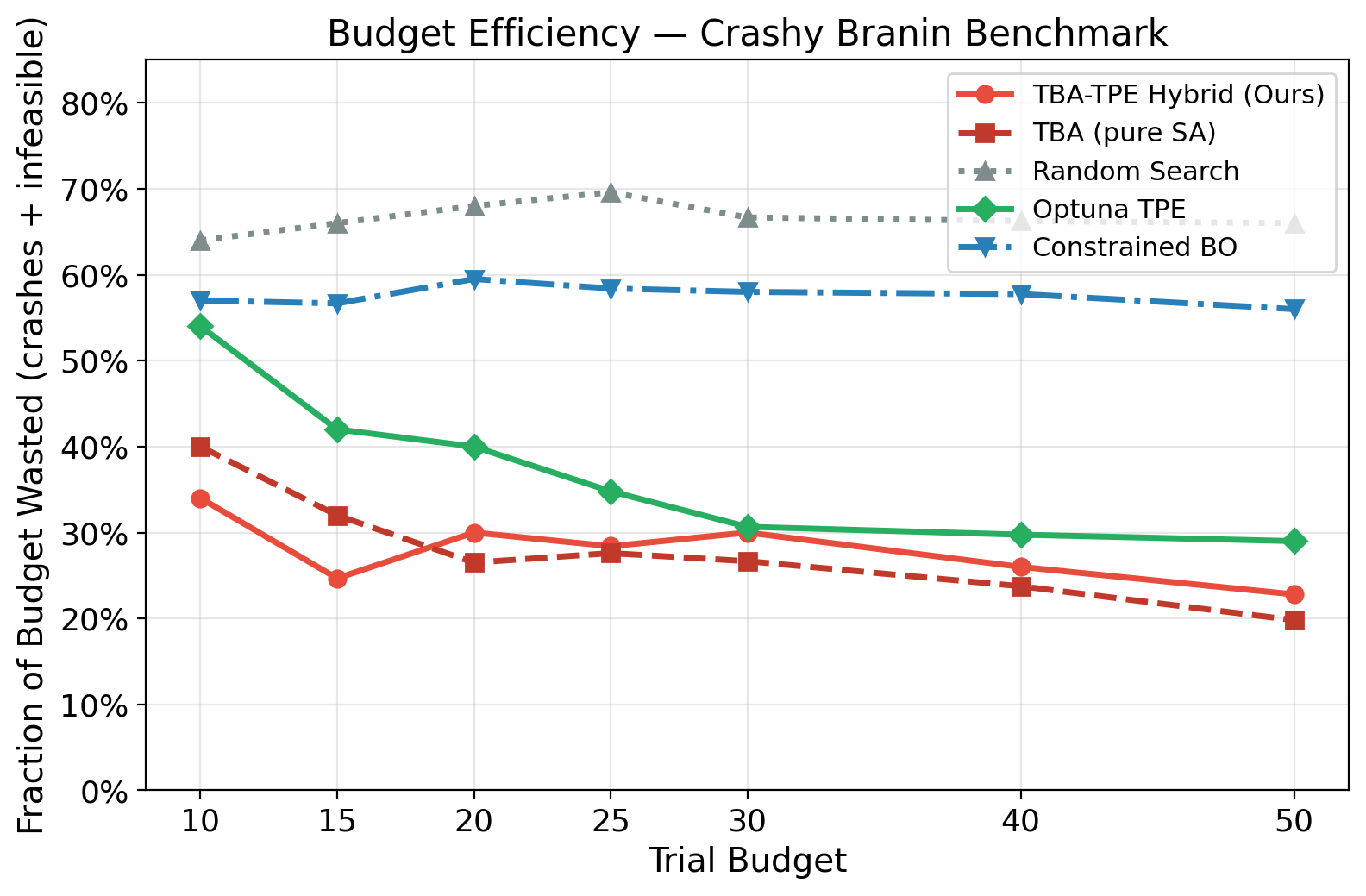}
\caption{Wasted budget fraction versus evaluation budget on Crashy Branin. The hybrid consistently wastes less budget than TPE across all budgets tested. Random search wastes 63--67\% regardless of budget.}
\label{fig:budget_wasted}
\end{figure}
 
\subsection{Ablation}
 
\begin{table}[t]
\centering
\caption{Ablation on Crashy Branin at $B=30$.}
\label{tab:ablation}
\begin{tabular}{lcc}
\toprule
\textbf{Variant} & \textbf{Best Objective} & \textbf{Wasted \%} \\
\midrule
TBA full (SA + all components) & $-2.93 \pm 2.37$ & 27\% \\
\quad w/o no-late-restarts & $-3.62 \pm 2.58$ & 30\% \\
\quad w/o decaying snap-back & $-3.45 \pm 2.71$ & 29\% \\
\quad w/o adaptive $p_s$ & $-3.21 \pm 2.50$ & 28\% \\
\TBATPE{} hybrid & $\mathbf{-2.05 \pm 0.71}$ & 30\% \\
\bottomrule
\end{tabular}
\end{table}
 
The ablation indicates that restricting restarts to the first 70\% of budget is the most impactful component among the pure-\TBA{} modifications, while the TPE handoff provides the largest overall gain over pure SA with substantially reduced variance.
 
\subsection{Real Deployment: Edge-Tight Scenario (RTX~5080)}
 
Table~\ref{tab:edge_tight} reports results under tight edge deployment constraints on the primary hardware target.
 
\begin{table}[t]
\centering
\caption{Edge-tight deployment on RTX~5080: latency$_{p95}\le 20$ms, memory $\le 512$MB, budget=25, 10 seeds.}
\label{tab:edge_tight}
\begin{tabular}{lccc}
\toprule
\textbf{Optimizer} & \textbf{Accuracy (mean $\pm$ std)} & \textbf{Wasted \%} & \textbf{Feasible seeds} \\
\midrule
Random Search & $0.762 \pm 0.008$ & 74\% & 10/10 \\
\TBATPE{} Hybrid & $0.761 \pm 0.010$ & 42\% & 10/10 \\
TBA (pure SA) & $0.752 \pm 0.019$ & 30\% & 10/10 \\
Optuna TPE & $0.752 \pm 0.010$ & 36\% & 10/10 \\
\bottomrule
\end{tabular}
\end{table}
 
The aggregate accuracies are close, and the hybrid's edge-tight mean accuracy (0.761) is essentially tied with random search (0.762). The hybrid's advantage is not a large mean-accuracy gain but rather improved discovery behavior and budget efficiency (Figure~\ref{fig:edge_tight}). The more informative result is model-family discovery.
 
\begin{figure}[t]
\centering
\includegraphics[width=0.85\textwidth]{fig3_deployment_edge_tight.png}
\caption{Edge-tight deployment results on RTX~5080: accuracy and wasted budget fraction by optimizer. The hybrid achieves accuracy comparable to random search while wasting roughly half the budget.}
\label{fig:edge_tight}
\end{figure}
 
Table~\ref{tab:discovery} shows that cold-start TPE discovers \texttt{vit\_tiny}, the best constrained model family in this setting (76.6\% accuracy), in only 3 out of 10 seeds. In the remaining seeds, TPE converges to \texttt{resnet50} (accuracy 0.746) or \texttt{efficientnet\_b0} (accuracy 0.742) and exploits them without evaluating \texttt{vit\_tiny}. This is the premature exploitation phenomenon described in Remark~\ref{rem:premature}.
 
\begin{table}[t]
\centering
\caption{Seeds (out of 10) discovering \texttt{vit\_tiny}, the globally best architecture under edge-tight constraints on the RTX~5080.}
\label{tab:discovery}
\begin{tabular}{lc}
\toprule
\textbf{Optimizer} & \textbf{Discovery rate} \\
\midrule
\TBATPE{} Hybrid & \textbf{8 / 10} \\
Random Search & 8 / 10 \\
TBA (pure SA) & 6 / 10 \\
Optuna TPE & 3 / 10 \\
\bottomrule
\end{tabular}
\end{table}
 
This is the central empirical finding. The hybrid avoids the \texttt{resnet50} trap because Phase~1 explores broadly enough to encounter \texttt{vit\_tiny} before the exploitation phase narrows the search. Random search also discovers \texttt{vit\_tiny} frequently, but it wastes 74\% of its budget on invalid trials, nearly double the hybrid's 42\%. Our primary practical objective is not universal objective dominance, but improved structural discovery under tight budgets with lower wasted evaluation cost.
 
The per-seed model discovery table (Table~\ref{tab:per_seed}) reveals the behavioral difference in detail. TPE converges to \texttt{resnet50} in 7 of 10 seeds without ever discovering \texttt{vit\_tiny}, while the hybrid's structural exploration during Phase~1 reaches \texttt{vit\_tiny} in 8 of 10 seeds before exploitation begins.
 
\begin{table}[t]
\centering
\caption{Best model family discovered per seed under edge-tight constraints on RTX~5080. \texttt{vit} = ViT-Tiny (0.766), \texttt{r50} = ResNet-50 (0.746), \texttt{r18} = ResNet-18 (0.720), \texttt{eff} = EfficientNet-B0 (0.742), \texttt{mob} = MobileNet-V2 (0.718).}
\label{tab:per_seed}
\begin{tabular}{c|cccc}
\toprule
Seed & \TBATPE{} & TBA (SA) & Random & TPE \\
\midrule
0 & vit & vit & vit & vit \\
1 & vit & r50 & r50 & r50 \\
2 & vit & r18 & vit & r50 \\
3 & vit & mob & vit & vit \\
4 & eff & vit & vit & r50 \\
5 & vit & vit & vit & r50 \\
6 & vit & vit & vit & eff \\
7 & vit & vit & vit & r50 \\
8 & eff & eff & r50 & r50 \\
9 & vit & vit & vit & vit \\
\midrule
vit rate & 8/10 & 6/10 & 8/10 & 3/10 \\
\bottomrule
\end{tabular}
\end{table}
 
\begin{figure}[t]
\centering
\includegraphics[width=0.85\textwidth]{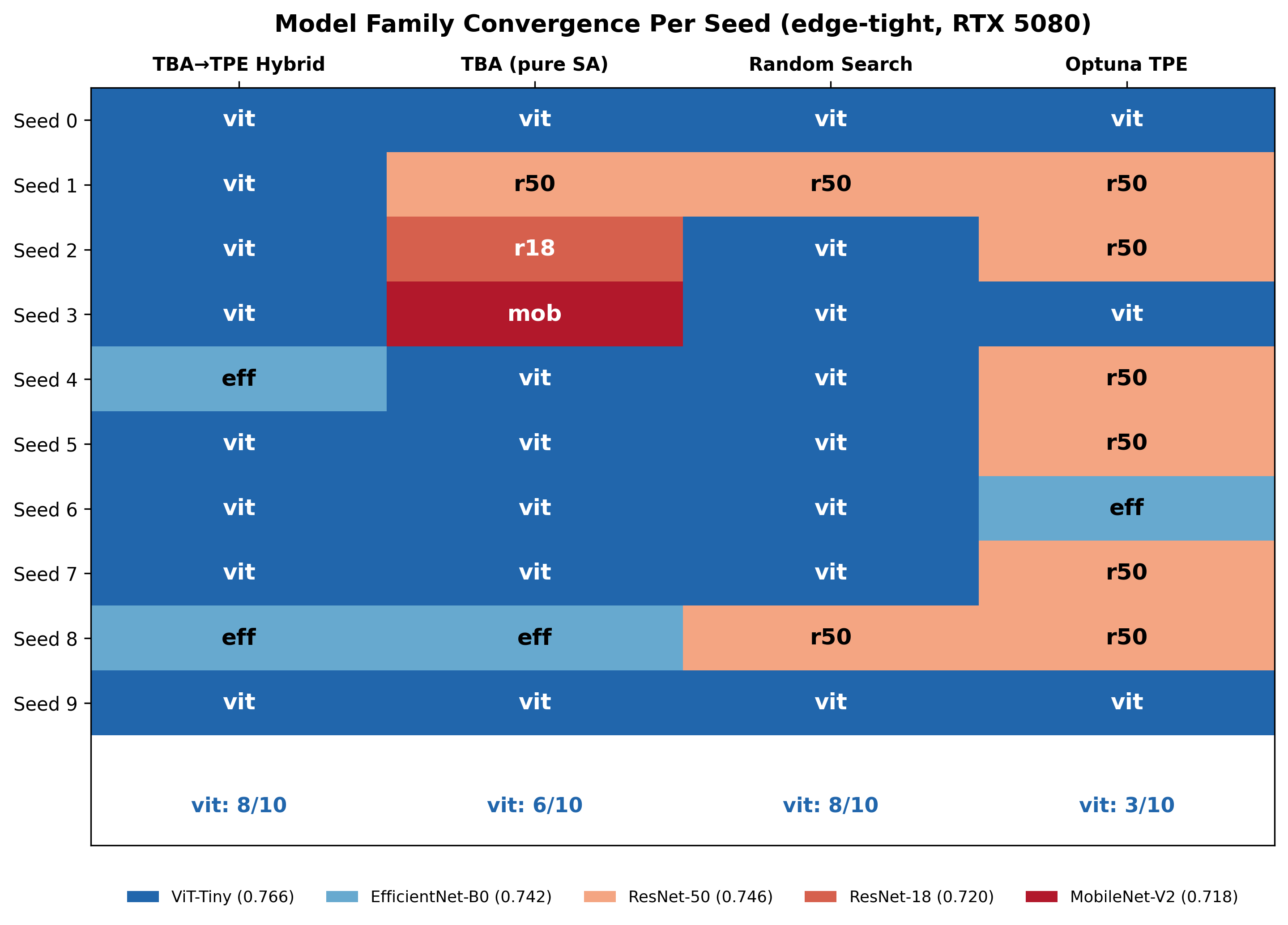}
\caption{Model family convergence per seed under edge-tight constraints on RTX~5080. Each cell shows which model family each optimizer converged to. TPE locks onto \texttt{resnet50} in 7 of 10 seeds, while the hybrid discovers \texttt{vit\_tiny} in 8 of 10 seeds.}
\label{fig:convergence}
\end{figure}
 
\subsection{Multi-Hardware Generalization}
\label{sec:multi_hw}
 
To test whether the hybrid's advantage generalizes across hardware targets with different search-space characteristics, we repeated the edge-tight experiment on four additional GPUs: NVIDIA H100, A100, L4, and T4 (5 seeds each). Each GPU induces a different \emph{effective hostility level} because hardware speed determines which configurations satisfy the 20ms latency constraint. Faster GPUs admit more feasible configurations, reducing the crash-and-infeasibility rate; slower GPUs tighten the feasibility boundary.
 
Table~\ref{tab:multi_hw_discovery} reports the \texttt{vit\_tiny} discovery rate across all five GPUs. Table~\ref{tab:multi_hw_summary} reports mean accuracy and wasted budget fraction.
 
\begin{table}[t]
\centering
\caption{Seeds discovering \texttt{vit\_tiny} under edge-tight constraints across five GPU targets. RTX~5080 uses 10 seeds; all others use 5 seeds.}
\label{tab:multi_hw_discovery}
\begin{tabular}{lccccc}
\toprule
\textbf{Optimizer} & \textbf{H100} & \textbf{A100} & \textbf{RTX 5080} & \textbf{L4} & \textbf{T4} \\
\midrule
\TBATPE{} Hybrid & \textbf{5/5} & \textbf{4/5} & \textbf{8/10} & \textbf{4/5} & 3/5 \\
TBA (pure SA) & 3/5 & 2/5 & 6/10 & 1/5 & 2/5 \\
Random Search & \textbf{5/5} & \textbf{5/5} & 8/10 & \textbf{4/5} & \textbf{4/5} \\
Optuna TPE & 3/5 & \textbf{4/5} & 3/10 & 3/5 & 3/5 \\
\bottomrule
\end{tabular}
\end{table}
 
\begin{table}[t]
\centering
\caption{Mean accuracy and wasted budget fraction under edge-tight constraints across five GPU targets.}
\label{tab:multi_hw_summary}
\begin{tabular}{l|cc|cc|cc|cc|cc}
\toprule
& \multicolumn{2}{c|}{\textbf{H100}} & \multicolumn{2}{c|}{\textbf{A100}} & \multicolumn{2}{c|}{\textbf{RTX 5080}} & \multicolumn{2}{c|}{\textbf{L4}} & \multicolumn{2}{c}{\textbf{T4}} \\
\textbf{Optimizer} & Acc & $W$ & Acc & $W$ & Acc & $W$ & Acc & $W$ & Acc & $W$ \\
\midrule
\TBATPE{} & .766 & 28\% & .762 & 32\% & .761 & 42\% & .761 & 32\% & .758 & 36\% \\
TBA (SA) & .752 & 25\% & .743 & 31\% & .752 & 30\% & .738 & 27\% & .739 & 33\% \\
Random & .766 & 63\% & .766 & 66\% & .762 & 74\% & .762 & 67\% & .760 & 72\% \\
TPE & .758 & 32\% & .762 & 33\% & .752 & 36\% & .758 & 34\% & .758 & 38\% \\
\bottomrule
\end{tabular}
\end{table}
 
\begin{figure}[t]
\centering
\includegraphics[width=\textwidth]{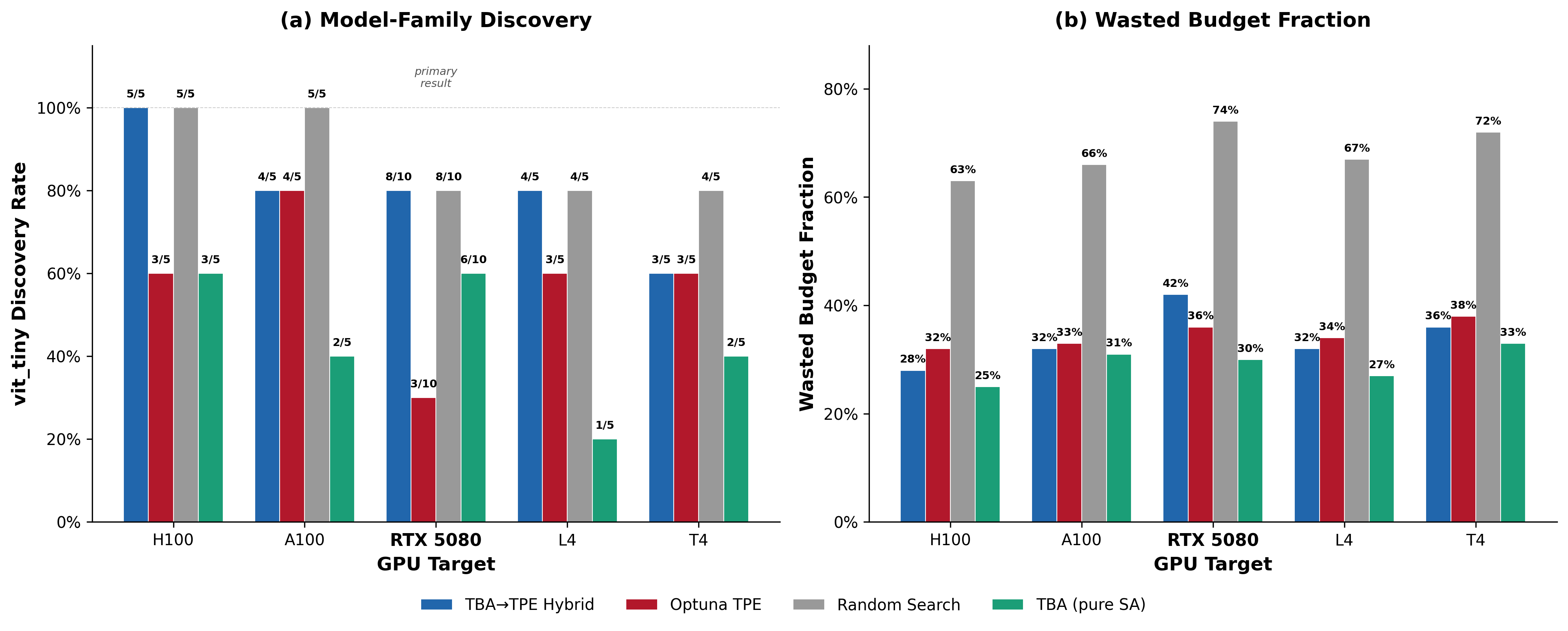}
\caption{Discovery rate and wasted budget fraction across five GPU targets. The hybrid achieves high discovery with moderate waste, while TPE achieves lower waste but misses the best model family in many seeds. RTX~5080 uses 10 seeds; the other GPUs use 5 seeds. No other optimizer occupies the high-discovery, low-waste quadrant.}
\label{fig:discovery_waste}
\end{figure}

Four findings emerge from the multi-hardware experiments. Figure~\ref{fig:discovery_waste} visualizes the central trade-off: the hybrid (dark blue) achieves discovery rates comparable to random search (gray) while wasting far less budget, and it never falls below TPE (red) on discovery despite maintaining comparable or lower waste.
 
\paragraph{The hybrid matches or exceeds TPE on discovery.}
The hybrid discovers \texttt{vit\_tiny} at a rate equal to or higher than TPE on all five hardware targets: 5/5 vs 3/5 on H100, 4/5 vs 4/5 on A100, 8/10 vs 3/10 on RTX~5080, 4/5 vs 3/5 on L4, and 3/5 vs 3/5 on T4. The gains are largest on H100 and RTX~5080; on A100 and T4, the methods are comparable.
 
\paragraph{The hybrid achieves lower waste than TPE on four of five GPUs.}
The hybrid wastes 28--42\% of budget compared to TPE's 32--38\%. On the H100, A100, L4, and T4, the hybrid's waste is lower than or equal to TPE's. On the RTX~5080, the hybrid's waste is higher (42\% vs 36\%) because it explores more broadly---but this broader exploration is precisely what drives its superior discovery rate (8/10 vs 3/10).
 
\paragraph{The hybrid's advantage is most visible on the H100 and RTX~5080.}
On the H100, the fastest GPU, the hybrid achieves perfect 5/5 discovery with only 28\% waste---matching random search's discovery while wasting less than half the budget. On the RTX~5080, with 10 seeds, the hybrid's advantage is most visible: 8/10 vs 3/10 for TPE. The H100 result may seem surprising---why does a fast GPU with many feasible configs still show a gap? The answer is that TPE's failure mode is \emph{model selection}, not latency: in seeds 2 and 4, TPE locks onto \texttt{resnet50} early and never evaluates \texttt{vit\_tiny}, regardless of hardware speed.
 
\paragraph{The T4 confirms the v2 improvements.}
In an earlier codebase without trial timeouts and subspace blacklisting, the hybrid discovered \texttt{vit\_tiny} in only 2/5 T4 seeds, below TPE's 3/5. The exploration phase became trapped in \texttt{int8\_dynamic} variants that consumed 500ms+ per trial. With the v2 improvements, the hybrid now matches TPE at 3/5 on T4 while reducing waste (36\% vs 38\%). Early stopping saved 7--47 seconds per seed, and subspace blacklisting prevented repeated proposals of \texttt{int8\_dynamic} after consecutive failures. The T4 remains the most hostile environment, but it is no longer a failure case for the hybrid.
 
\subsection{The Role of Random Search}
 
An important observation is that random search matches the hybrid's \texttt{vit\_tiny} discovery rate (8/10) while wasting 74\% of budget. This is diagnostically useful. It confirms that \texttt{vit\_tiny} is not hidden in an obscure corner of the space---the problem is not reachability but \emph{search dynamics}. TPE's model-guided policy steers away from that family after committing early to \texttt{resnet50}. The hybrid preserves much of the structural coverage of broad random exploration while avoiding its extreme inefficiency.
 
\section{Analysis}
\label{sec:analysis}
 
\subsection{Why Does TPE Miss the Best Model?}
 
TPE's failure to discover \texttt{vit\_tiny} in 7/10 RTX~5080 seeds and 2/5 H100 seeds can be explained by the interaction between crash rate and exploration depth. TPE uses an initial random design before switching to model-guided sampling. In a space where 65--80\% of configurations crash or violate constraints, those initial trials may yield only 2--4 successful evaluations, potentially all from the same model family. Once the density estimator is fitted to this narrow sample, exploitation reinforces the same family rather than expanding structural coverage.
 
Using the formula from Remark~\ref{rem:premature} with $K=5$, $\rho_{\text{crash}}=0.6$, and $n_0=10$: $\Prob \approx 0.57$. This predicts that roughly 4--6 out of 10 seeds would discover any particular family, which is directionally consistent with the observed TPE discovery rates across hardware.
 
By contrast, the hybrid uses Phase~1 to widen structural coverage before handing off to TPE. This does not guarantee discovery of every promising family, but it raises the probability that the best family enters the candidate set prior to exploitation.
 
\subsection{The Trade-Off Between Discovery and Waste}
 
Across all five GPUs, the hybrid consistently occupies a distinct position in the discovery--waste trade-off:
\begin{itemize}[leftmargin=*]
\item Random search achieves high discovery but extreme waste (63--74\%).
\item TPE achieves lower waste (32--38\%) but misses the best model family in many seeds.
\item Pure SA achieves the lowest waste (25--33\%) but often converges to the wrong model family (resnet18, mobilenet\_v2).
\item The hybrid achieves high discovery with moderate waste (28--42\%)---the best overall trade-off.
\end{itemize}
No other optimizer achieves both high discovery \emph{and} low waste simultaneously. This is the practical value of the feasible-first exploration design.
 
\subsection{Exploration--Exploitation Boundary}
 
The multi-hardware experiments suggest that the hybrid's advantage is present across all hostility levels, though its magnitude varies. On fast GPUs (H100), the hybrid's broader Phase~1 exploration finds \texttt{vit\_tiny} in every seed because the feasible region is large (batch sizes 1--32 all meet the 20ms constraint). On slow GPUs (T4), the feasible region is narrow and exploration traps are more likely, but trial timeouts and subspace blacklisting mitigate the worst cases. The hybrid's advantage is most visible on mid-range hardware (RTX~5080, L4) where the space is hostile enough that TPE struggles but not so hostile that even deliberate exploration becomes trapped.
 
\section{Discussion}
\label{sec:discussion}
 
The results support the paper's central claim with appropriate nuance. The hybrid is not universally better than TPE on every metric. Rather, it changes the exploration--exploitation trade-off in settings where the early trial budget is too small, and the invalidity rate too high, for cold-start model-guided search to form a sufficiently broad view of the space.
 
H1 (structural coverage) is supported by the discovery-rate analysis: the hybrid's structural mutations during Phase~1 produce broader model-family coverage than TPE's cold-start random sampling in crash-heavy spaces. The effect is present across all five GPUs, with the hybrid matching or exceeding TPE's discovery rate on every hardware target, though the margin varies from substantial (RTX~5080: 8/10 vs 3/10) to negligible (T4: 3/5 vs 3/5).
 
H2 (invalidity changes the trade-off) is supported by the contrasting results across hardware targets: the hybrid's discovery advantage is largest on hardware where the crash-and-infeasibility rate is high enough to impair TPE's initial sampling (RTX~5080: 8/10 vs 3/10) but present even on the fastest hardware (H100: 5/5 vs 3/5).
 
H3 (warm-start helps) is supported by the ablation: \TBATPE{} achieves $-2.05 \pm 0.71$ versus TBA pure SA's $-2.93 \pm 2.37$ at $B=30$, with both better mean and substantially lower variance. The multi-hardware discovery rates further confirm this: pure SA discovers \texttt{vit\_tiny} at lower rates than the hybrid on all tested GPUs (e.g., 3/5 vs 5/5 on H100, 1/5 vs 4/5 on L4, 6/10 vs 8/10 on RTX~5080).
 
A secondary implication concerns evaluation metrics. In deployment optimization, best feasible objective alone is insufficient. The hybrid's advantage manifests primarily in budget efficiency and discovery consistency, not in mean objective superiority. Trial waste is operationally meaningful because each evaluation may require model export, runtime compilation, repeated timing, and memory measurement. A method that obtains comparable feasible performance with substantially lower waste can be preferable in practice.
 
\section{Limitations and Threats to Validity}
\label{sec:limitations}
 
Several factors limit the scope of the present conclusions.
 
\paragraph{Hardware coverage.}
While multi-hardware experiments span five GPU targets from datacenter (H100, A100) to edge (T4) and laptop (RTX~5080), all experiments use the same software stack and model set. Backend behavior and quantization support may differ on non-NVIDIA hardware or with different compiler versions.
 
\paragraph{Single deployment domain.}
The benchmark focuses on image classification deployment with a single scenario (edge-tight). Extensions to NLP serving, speech inference, multi-model pipelines \citep{zaharia2024compound}, and additional constraint profiles remain future work.
 
\paragraph{Benchmark scale.}
The deployment space contains roughly 2000 structural combinations with five model families. Larger spaces with more model families may amplify the premature exploitation effect, but they may also challenge the exploration capacity of Phase~1.
 
\paragraph{Statistical scope.}
Ten seeds are used for the primary RTX~5080 experiments and five seeds for multi-hardware experiments. Larger seed counts would strengthen the conclusions, particularly for the T4 and L4 where differences of 1 seed (e.g., 3/5 vs 4/5) are not statistically significant.
 
\paragraph{Aggregate accuracy parity.}
The hybrid's edge-tight mean accuracy (0.761) is essentially tied with random search (0.762). Its advantage manifests primarily in discovery behavior and budget efficiency rather than large gains in average objective.
 
\paragraph{Residual exploration traps.}
While trial timeouts and subspace blacklisting substantially mitigate the T4 failure mode (improving from 2/5 to 3/5), the hybrid still does not achieve the same discovery rate on T4 as on faster hardware. On the A100, the hybrid also missed \texttt{vit\_tiny} in seed~3 due to the SA phase becoming trapped in \texttt{int8\_dynamic} variants that were slow but not slow enough to trigger the $5\times$ early-stopping threshold.
 
\paragraph{Synthetic--real gap.}
The synthetic benchmarks reproduce structural invalidity and hidden crash regions but do not replicate every detail of real deployment failures.
 
\section{Conclusion}
\label{sec:conclusion}
 
This paper studied deployment optimization as a budget-constrained search problem in a partially invalid hierarchical space. The main result is not that simulated annealing outperforms mature model-guided optimizers in general. Instead, the evidence supports a narrower and more useful conclusion: when invalidity is high and budgets are small, \emph{the quality of early exploration determines the effectiveness of later exploitation}. Warm-starting TPE with feasible-first exploration improves model-family discovery under tight constraints and reduces wasted budget relative to cold-start TPE.
 
Multi-hardware experiments across five GPU targets confirm that the hybrid never underperforms TPE on discovery, with the clearest gains on H100 (5/5 vs 3/5) and RTX~5080 (8/10 vs 3/10). On A100 and L4, the hybrid matches or slightly exceeds TPE (4/5 vs 4/5 and 4/5 vs 3/5, respectively). On the most hostile hardware, after the addition of trial timeouts and subspace blacklisting, the hybrid matches TPE with lower waste (T4: 3/5 vs 3/5, 36\% vs 38\% waste).
 
Three directions appear especially promising for future work. First, the handoff criterion could be made adaptive to observed structural diversity rather than simple counts, requiring minimum model-family coverage before initiating TPE. Second, hardware-transfer experiments could test whether feasibility maps learned on one device reduce search cost on another. Third, the benchmark can be extended beyond image classification to include additional deployment scenarios, model families, and constraint profiles.
 
The benchmark suite and optimizer implementation are available at \url{https://github.com/Chrislysen/Constrained-ML-Deployment}.
 
\appendix
 
\section{Additional Method Detail}
\label{app:method}
 
\subsection{Adaptive Structural Mutation Schedule}
 
The structural mutation probability is decayed linearly:
\begin{equation}
p_s(n) = 0.5 - 0.35 \cdot \frac{n - n_{\text{init}}}{B - n_{\text{init}}},
\end{equation}
where $n_{\text{init}}$ is the number of initial random trials.
 
\subsection{Snap-Back and Elite Restart}
 
The pure-\TBA{} variant uses two stabilization mechanisms: (i) elite restart to the 2nd or 3rd best feasible state during early search, and (ii) decaying snap-back to the best feasible state after accepting a worse move. These mechanisms were retained because they improved the ablation metrics on Crashy Branin (Table~\ref{tab:ablation}).
 
\subsection{Subspace Blacklisting Detail}
 
The subspace tracker maintains a dictionary mapping each (categorical variable, value) pair to a consecutive failure count. On each trial outcome:
\begin{itemize}[leftmargin=*]
\item \textbf{Crash or infeasible}: increment failure count for all categorical values in the configuration. If count reaches 3, blacklist the value.
\item \textbf{Success (feasible)}: reset failure count to 0 for all categorical values in the configuration; remove any active blacklist.
\item \textbf{Cooldown}: after 8 trials since blacklisting, automatically remove the blacklist regardless of outcomes.
\end{itemize}
The safety valve ensures that for each categorical variable, at least one value remains unblacklisted at all times.
 
\section{Computational Cost}
\label{app:compute}
 
\begin{table}[H]
\centering
\caption{Wall-clock time for complete experimental evaluation.}
\label{tab:compute}
\begin{tabular}{lr}
\toprule
\textbf{Experiment} & \textbf{Wall-clock} \\
\midrule
Synthetic budget sweep (4 opt. $\times$ 7 budgets $\times$ 10 seeds) & $\sim$15 min \\
Edge-tight deployment RTX~5080 (4 opt. $\times$ 10 seeds $\times$ 25 trials) & $\sim$120 min \\
Multi-hardware edge-tight (4 opt. $\times$ 5 seeds $\times$ 25 trials $\times$ 4 GPUs) & $\sim$6 hours \\
\midrule
\textbf{Total} & $\sim$8.5 hours \\
\bottomrule
\end{tabular}
\end{table}
 
Primary experiments were conducted on a single NVIDIA RTX~5080 laptop GPU. Multi-hardware experiments were conducted on NVIDIA H100~80GB, A100~40GB, L4~24GB, and T4~16GB GPUs via Google Colab.
 
\section{Reproducibility Checklist}
\label{app:repro}
 
\begin{enumerate}[leftmargin=*]
\item All experiments across all hardware targets use the same codebase (v2-improvements branch, commit \texttt{7ea93e2}).
\item Primary experiments used 10 random seeds (0--9); multi-hardware experiments used 5 seeds (0--4).
\item Crashed and early-stopped trials were logged rather than discarded.
\item The search-space specification, benchmark definitions, and experiment scripts are available in the public repository.
\item The reported tables are generated from saved JSONL result files.
\item Conditional parameters are exposed identically to all optimizers.
\item Trial timeouts are applied at the benchmark level and affect all optimizers equally.
\item Multi-hardware experiments were conducted in single clean sessions per GPU to avoid cross-session variability.
\end{enumerate}
 
\bibliographystyle{plainnat}

\end{document}